\pdfoutput=1

\documentclass[11pt]{article}
\usepackage{amsmath} 
\usepackage[final]{acl}
\usepackage{mdframed}
\usepackage{array}
\usepackage{graphicx} 
\usepackage{float}
\usepackage{enumitem} 
\usepackage{expex}

\usepackage{times}
\usepackage{latexsym}

\usepackage[T1]{fontenc}

\usepackage[utf8]{inputenc}

\usepackage{microtype}

\usepackage{inconsolata}

\title{Exploring Factual Entailment with NLI: A News Media Study}

\author{Guy Mor-Lan \\
  Hebrew University of Jerusalem \\
  \texttt{guy.mor@mail.huji.ac.il} \\\And
  Effi Levi \\
  Hebrew University of Jerusalem \\
 \texttt{efle@cs.huji.ac.il} \\
}

\begin{document}
\maketitle
\begin{abstract}
We explore the relationship between factuality and Natural Language Inference (NLI) by introducing FactRel -- a novel annotation scheme that models \textit{factual} rather than \textit{textual} entailment, and  use it to annotate a dataset of naturally occurring sentences from news articles. Our analysis shows that 84\% of factually supporting pairs and 63\% of factually undermining pairs do not amount to NLI entailment or contradiction, respectively, suggesting that factual relationships are more apt for analyzing media discourse. We experiment with models for pairwise classification on the new dataset, and find that in some cases, generating synthetic data with GPT-4 on the basis of the annotated dataset can improve performance. Surprisingly, few-shot learning with GPT-4 yields strong results on par with medium LMs (DeBERTa) trained on the labelled dataset. We hypothesize that these results indicate the fundamental dependence of this task on both world knowledge and advanced reasoning abilities.

\end{abstract}

\section{Introduction}

In recent years, the concept of factuality in news media has garnered increasing attention. Studies increasingly examine the relation between facts - as presented in news coverage - and phenomena such as political polarization, misinformation and fake news \citep{roy-goldwasser-2020-weakly, levy2021social, bakshy2015exposure, garimella2021political}. 
As a result, the ability to model factual relations between claims becomes increasingly important. This has led to a line of work on automated fact-checking, which involves textual pipelines for detecting and evaluating factual claims \citep{zeng2021automated}.

In automatic fact-checking, fact verification is predominantly addressed via the Natural Language Inference (NLI) task, also known as Recognizing Textual Entailment (RTE) \citep{zeng2021automated, arana-catania-etal-2022-natural, nie2018combining, sathe-etal-2020-automated}, which has been used for decades for evaluating natural language understanding capabilities \citep{poliak-2020-survey}. NLI is traditionally formulated as a categorical classification task between a premise \( p \) and a hypothesis \( h \), where \( p \) can either contradict, entail or be neutral with respect to \( h \). Large NLI datasets such as SNLI and MNLI \citep{bowman-etal-2015-large, williams-etal-2018-broad} have become highly popular, leading NLI to be adapted to various uses such as zero-shot classification \citep{yin-etal-2019-benchmarking} and semantic similarity \citep{reimers-gurevych-2019-sentence}. In fact verification, NLI is used to evaluate the relations between a candidate fact and trusted pieces of evidence \citep{zeng2021automated}.



However, the adequacy of NLI for analyzing factual relationships in news media is hindered by 
two primary reasons, relating to the nature of the task as well as to the characteristics of commonly used NLI datasets. First, large NLI datasets such SNLI and MNLI define the pairwise relationship in 
terms 
of necessity of meaning
\citep{bowman-etal-2015-large, williams-etal-2018-broad}. 
Thus, in MNLI an entailment is defined to be the case whereby a hypothesis ``is necessarily true or appropriate whenever the premise is true'', and similarly a contradiction is when the hypothesis ``is necessarily false or inappropriate whenever the premise is true'' \citep{williams-etal-2018-broad}. However, 
these types of
relationships may be too restrictive
for the analysis of media discourse, where explicit 
contradictions and entailments are likely to be rare, as such discourse tends take place in the margins of plausibility.

Secondly, texts in popular NLI datasets considerably differ from news texts. While sentences in NLI datasets tend to be short, simple, highly generic and convey a single idea or statement, media sentences tend to be longer, more complex, more specific and convey multiple pieces of information. 




A common feature of NLI datasets such as RTE, SNLI and MNLI is that while premises are naturally occurring texts, the hypotheses are specifically written to correspond to the categories \citep{chatzikyriakidis-etal-2017-overview, williams-etal-2018-broad}. While this method is effective in generating large amounts of data, constructed hypotheses are likely to express a simple relationship to the premise and thus not resemble pairs of naturally occurring sentences. Additionally, \citet{chatzikyriakidis-etal-2017-overview} notes that these datasets feature strictly logical relationships and stresses the need for datasets capturing other sorts of inferential relationships.


In this work, we set out to examine the relationship between NLI and textual factuality. For this purpose, we have developed a novel annotation scheme that expresses \textit{factual} rather than \textit{textual} entailment, encoding each pair of sentences with the relation of factual support, factual undermining, or neither.  We have annotated a new dataset of naturally occurring sentence pairs from news media using both our factual entailment scheme and NLI, enabling a comparison of the schemes on news media. We also check the ability of recent generative
LLMs (GPT-4) to 
generate such pairs correctly. 
We end with a set of experiments that demonstrate the ability to learn the factual entailment task using fine-tuned models as well as generative LLMs, and draw conclusions regarding the task's relation to real world knowledge in comparison to NLI. Overall, we analyze differences between NLI and factual entailment in their scope, relevance to news text and dependence on world knowledge, and show potential for new ways to model factual relations.


\section{Factual Entailment} 

For the purpose of exploring the relationship between factual relations and textual entailment, we have developed FactRel, a novel annotation scheme encoding the \textit{factual entailment} between pairs of sentences. Similarly to NLI, FactRel is a 3-category pairwise classification task. Given a premise \( p \) and a hypothesis $h$, $p$ can either factually support $h$ (\textit{SUPPORT}), factually undermine $h$ (\textit{UNDERMINING}), or be factually neutral w.r.t $h$ (\textit{NEUTRAL}). $p$ is said to factually support $h$ when $p$ being true would make $h$ more plausible or likely to be true, compared to a situation in which the truth value of $p$ is unknown. $p$ is said to factually undermine $h$ when $p$ being true would make $h$ less plausible or likely to be true, compared to a situation in which the truth value of $p$ is unknown. Finally, $p$ is said to be factually neutral w.r.t to $p$ when $p$'s truth has no bearing on the plausibility of $h$, and the likelihood of $h$ would not change if $p$ was known to be either true or false.

While both NLI and FactRel encode a ternary entailment relation between pairs of sentences, the factual relation encoded by FactRel is quite different from the one encoded by NLI. For example, consider the following pair of sentences:
\pex
\a[label=\textit{p}] ``You can't run a festival or you can't run a nightclub or a live-music gig with social distancing,'' Lord said.
\a[label=\textit{h}] Peter Marks, the CEO of Rekom, Britain's largest specialist late-night bar operator, told Insider the company's venues were set to open on June 21 ``without COVID measures.''
\xe
The above example exhibits a relation of factual \textit{SUPPORT} while its NLI label is \textit{NEUTRAL}. The hypothesis matches the premise and exemplifies it, but the premise does not necessitate the hypothesis.

A parallel example can be observed in the following pair of sentences:
\pex
\a[label=\textit{p}] FILE – In this April 12, 2021 file photo, people queue outside a Hermes store in Mayfair in London.
\a[label=\textit{h}] Sales of luxury apparel, jewelry, leather goods and beauty products plunged to 217 billion euros in the pandemic year of 2020, from 281 billion euros in 2019, shedding six years of growth.
\xe
This example exhibits a relation of factual \textit{UNDERMINING} while its NLI label is \textit{NEUTRAL}. There is factual tension between the premise and hypothesis, as the premise can be considered a counter-example to the hypothesis, but it does not necessitate the hypothesis' falsity.

There are, however, cases in which the two schemes converge to the same relation. For example, 
\pex
\a[label=\textit{p}] Woman accused of attempted murder after driving into President Trump supporters in Southern California
\a[label=\textit{h}] The vast majority of those cases tallied by Weil involved motorists who ran into those demonstrating for causes aligned with the Black Lives Matter movement, Weil said.
\xe
This example is factually \textit{NEUTRAL}, and its NLI label is \textit{NEUTRAL} as well.


\section{Dataset}
\subsection{Construction}
The core dataset comprises 1,507 sentence pairs sampled from 211 news articles appearing in diverse English-language digital news outlets in the period 2020-2022. Pairs were sampled from the same news article in order to increase the likelihood of the pairs having a non-neutral relationship. The sentence pairs were independently labelled by two annotators -- one of the authors and a research assistant -- with a subset annotated by both for calculating inter-coder reliability (Table~\ref{tab:intercoder_reliability}). Annotators are instructed to categorize only non-negligible relations of support and undermining as such. Conflicts were resolved by committee consultation.

\begin{table}
\centering
\begin{tabular}{l|cc}
\hline
\textbf{Item} & \textbf{Agreement \%} & \textbf{Kappa} \\
\hline
\textbf{Factual Entailment} & 95.2\% & 0.93 \\
\textbf{NLI} & 95.2\% & 0.85 \\
\hline

\end{tabular}
\caption{Intercoder reliability for annotations of NLI and factual entailment, showing raw agreement rate and Cohen's Kappa.}
\label{tab:intercoder_reliability}
\end{table}

The core dataset is augmented by two additions. First, a subset of 500 sentence pairs from the MNLI dataset was annotated with factual entailment,
for the purpose of examining differences between the MNLI dataset and the proposed dataset. Secondly, a synthetic dataset was generated using GPT-4 on the basis of the training set split from the core dataset. Each sentence pair in the training set was sent to GPT-4 accompanied by an explanation of the factual relationship task, the annotated label for that pair, and the definition of the label. GPT-4 was asked to generate 10 diverse examples possessing the same label, modelled on the sentence pair from the annotated dataset (see appendix~\ref{app:synt_dataset} for prompts). Thus, the synthesized addendum is 10 times larger than the core training set and consists of 12,050 pairs. A subset of 500 GPT generated pairs was randomly sampled for manual validation, showing that in 98.4\% of the pairs the manual labelling is consistent with GPT.

\subsection{Analysis} 

In the core dataset, 93\% of sentence pairs are 
NLI-neutral, whereas a smaller share of 70\% are factually neutral (see Table~\ref{tab:crosstab_core}). This indicates that non-neutral factual relationships are significantly more common in news media than non-neutral 
NLI
relationships. In terms of length, we observe a significant difference between FactRel and NLI datasets -- the average number of tokens per sentence in FactRel is $20.2$, compared to $10.1$ and $15.01$ in the respective training splits of SNLI and MNLI.

The dual annotation of the dataset with factual entailment and NLI labels allows us to examine the relationship between the two. We examine the correlation between the labels utilizing Cramér's V association measure for discrete variables. While factual categories are strongly correlated with  
the 
categories in the MNLI subset (\( \phi_c = 0.72\)), the correlation is lower in the core dataset of news sentence pairs (\( \phi_c = 0.49\)). In the core dataset, 84\% of factually supporting pairs and 63\% of factually undermining pairs do not amount to
entailment or contradiction, respectively (Table~\ref{tab:crosstab_core}). In the MNLI subset, the numbers are respectively 32\% and 2\% (Table~\ref{tab:crosstab_mnli}). 
This discrepancy likely indicates how in real news discourse, factual relations are increasingly untangled from semantic necessity, compared to datasets such as MNLI which contain sentences specifically written to form relations of semantic necessity.

\begin{table}
\centering
\begin{tabular}{l|ccc}
\hline
\textbf{Factual / NLI} & \textbf{Contra.} & \textbf{Entail.} & \textbf{Neutral} \\
\hline
\textbf{Support}     & 0                      & 48                  & 245                \\
\textbf{Undermining} & 67                    & 0                    & 113                 \\
\textbf{Neutral}             & 0                     & 0                   & 1130                \\
\hline
\end{tabular}
\caption{Cross-tabulation between NLI and Factual Entailment, core dataset.}
\label{tab:crosstab_core}
\end{table}

\begin{table}
\centering
\begin{tabular}{l|ccc}
\hline
\textbf{Factual / NLI} & \textbf{Contra.} & \textbf{Entail.} & \textbf{Neutral} \\
\hline
\textbf{Support}     & 5                      & 155                  & 67                \\
\textbf{Undermining} & 174                    & 1                    & 2                 \\
\textbf{Neutral}             & 17                     & 10                   & 69                \\
\hline
\end{tabular}
\caption{Cross-tabulation between NLI and Factual Entailment, MNLI subset.}
\label{tab:crosstab_mnli}
\end{table}

\section{Experiments}


We 
tackle the task of factual entailment
with several types and sizes of language models.

\textbf{Baseline model.} As a simple baseline, we embed the premise and hypothesis using the UAE-Large-V1 encoder \citep{li2023angleoptimized} and calculate the cosine similarity between them, on which we train a decision tree with a max depth of 10.

\textbf{Zero shot and Few Shot (no training).} We use two models in a zero-shot setting. First, we utilize a state-of-the-art NLI model trained on many NLI datasets \citep{laurer2022less}. The NLI model, based on DeBERTa V3 large \citep{he2021debertav3}, was used as if the NLI categories are equivalent to FactRel categories (e.g., \textit{CONTRADICTION} equals \textit{UNDERMINING}). Second, we utilize GPT-4 in a zero-shot setting provided only with a description of the task and the categories. We additionaly use GPT-4 in a 3-shot setting, adding three example pairs, one for each category.

\textbf{Trained Models.} We fine-tune several encoder models: RoBERTa-base \citep{DBLP:journals/corr/abs-1907-11692}, DeBERTa V3 large \citep{he2021debertav3}, and DeBERTa V3 SOTA NLI checkpoint \citep{laurer2022less}. Training variants included training with class weights and utilizing focal loss. 
We also fine-tune GPT-3.5 using OpenAI's API with the recommended settings. All the models 
were tested using two types of training sets -- 
the core training set, and the augmented set with GPT-4 synthetic pairs added.
Full technical details of the training setup are laid out in appendix~\ref{app:training_setup}.

Macro-F1 results on the validation set for the baseline model, the stock NLI model and the top performing models 
are reported in Table~\ref{tab:model_performance} (see appendix~\ref{app:full_results} for full results). Table~\ref{tab:model_performance_augmented} examines the effect of adding synthetic data to the training set. Overall, the results show that while the task is learnable, it is not easy even for large pre-trained models. GPT-4 performs surprisingly well in both zero-shot and 3-shot settings, with GPT-4 3-shot being the most performant model, matching the Macro-F1 of finetuned DeBERTa with slightly better accuracy. The inclusion of synthetic data enhances the performance of the baseline model and DeBERTa-NLI, but decreases the performance of fine-tuned GPT-3.5. 

\begin{table}
\centering
\setlength{\tabcolsep}{3pt}
\begin{tabular}{lcc}
\hline
\textbf{Model}                                    & \textbf{F1}$_{\text{MAC}}$ & \textbf{ACC} \\ 
\hline
Baseline (Cosine similarity)                  & 0.38 & 0.61 \\ 
\hline
Stock NLI (no training)                                    & 0.54              & 0.72              \\ 
GPT-4 zero-shot                                   & 0.65              & 0.80              \\ 
GPT-4 3-shot                                      & \textbf{0.70}              & \textbf{0.81}              \\ 
\hline
Fine-tuned GPT-3.5                                & 0.69              & 0.78              \\ 
DeBERTa-NLI / Focal loss                      & 0.68              & 0.8              \\ 
\hline
\end{tabular}
\caption{Top performing models, core training set}
\label{tab:model_performance}
\end{table}

\begin{table}
\centering
\setlength{\tabcolsep}{3pt}
\begin{tabular}{lcc}
\hline
\textbf{Model}                                    & \textbf{F1}$_{\text{MAC}}$ & \textbf{ACC} \\ 
\hline
Baseline (Cosine similarity)                  & 0.44 & 0.63 \\ 
\hline
Fine-tuned GPT-3.5                                & 0.63              & 0.77              \\ 
DeBERTa-NLI / Focal loss            & \textbf{0.70}             & \textbf{0.79}              \\ 
\hline
\end{tabular}
\caption{Top trained models, augmented training set}
\label{tab:model_performance_augmented}
\end{table}

\section{Conclusion}

In this paper we explored the relationship between NLI and factual relations. For this purpose, we designed a new annotation scheme for factual entailment, FactRel; examined it in comparison to NLI on a sample of annotated pairs from news coverage; and examined the performance of various models on the task. We have shown that factual entailment relations are significantly more common in news articles in comparison to semantic entailment, thus underlining the shortcomings of NLI when applied to naturally occurring text.

We have also shown that GPT-4 performs better in a few-shot setting than smaller models trained on the entire training set. Moreover, GPT-4's performance even in a zero-shot setting is competitive with other models. The success of these LLMs, even with significantly less data, can give us insight on the challenge involved in the FactRel task and how it differs from NLI. 

NLI is a fundamentally semantic task, as determining whether \(p\) entails or contradicts \(h\) hinges on understanding the meaning of the words and concepts employed in both. Thus, if \(p\) semantically entails \(h\), then \(h\) itself must be included either explicitly or implicitly in \(p\) itself. The relations are therefore to be found in the meaning of the words.
Modelling factual relationships, on the other hand, also requires a significant amount of background knowledge on the referents of the words, a detailed world model, and nuanced reasoning abilities. Thus, in order to identify that the premise ``Twitter has locked Trump’s account for 12 hours, and required the removal of the tweets'' supports the hypothesis ``Facebook locked Trump’s account for 24 hours following two policy violations''
, it is required to not only understand the words and concepts, but to also 
be able to infer why a social network might lock one's account, and why such actions on two social networks are likely to co-occur.
It is thus hypothesized that LLMs that have broad world knowledge, and especially those that excel at reasoning such as GPT-4, are well placed for this task, and their world knowledge and reasoning capabilities can compensate for decreased exposure to training data.

Finally, the addition of synthesized data improves performance of the top medium size LM, showing that data synthesis can be successfully employed on this task. However, this improvement is not consistent for all configurations.

\section*{Limitations}
In line with NLI datasets, FactRel uses discrete classification labels. While the dataset distinguishes between semantic entailment and contradiction and (mere) factual support and undermining, it does not quantify the amount of support or undermining. However, the modelling of factual relationships can benefit from a probabilistic framework, which we leave to future research.

\section*{Acknowledgements}
This work was supported by the Israel Science Foundation (Grant no. 2501/22). We thank Hagar Kaminer for diligent research assistance as well as insightful comments and suggestions during the annotation process. We extend deep gratitude to Shaul R. Shenhav and Tamir Sheafer for their invaluable guidance, support and advice.

\bibliography{anthology,custom}

\appendix
\label{sec:appendix}

\newpage
\section{Synthetic Dataset}
\label{app:synt_dataset}

The synthetic component was created by generating 10 synthetic examples for each annotated sample in the training set, using GPT-4.

The following system prompt was used:

\begin{samepage}
\begin{mdframed}
\mdfsubtitle{SYSTEM PROMPT}
You are an advanced synthetic dataset generator.
\end{mdframed}
\end{samepage}

For factual support samples, the following prompt was used:

\begin{samepage}
\begin{mdframed}
\mdfsubtitle{FACTUAL SUPPORT PROMPT}
'Factual support' is a relationship between sentences A and B whereby A being true increases the likelihood of B being true.\\ \\
For example: \\
A: \{premise\} \\
B: \{hypothesis\} \\
\\
Generate 10 more pairs of sentences with a factual support relationship.
The sentences should be diverse and reflect the type of real life sentences normally found in news discourse. 
The sentences should resemble the provided example but should also vary. Like the provided example, the generated samples should not be overly simple. Each sentence pair should be separated with two newlines. \\ \\
Within each pair, the sentences should be separated with a single newline. Each sentence should start with 'A: ' or 'B: '. Apart from that do not generate any other output.
\end{mdframed}
\end{samepage}

For factual undermining samples, the following prompt was used:

\begin{mdframed}
\mdfsubtitle{FACTUAL UNDERMINING PROMPT}
'Factual undermining' is a relationship between sentences A and B whereby A being true decreases the likelihood of B being true.\\ \\
For example: \\
A: \{premise\} \\
B: \{hypothesis\} \\
\\
Generate 10 more pairs of sentences with a factual undermining relationship.
The sentences should be diverse and reflect the type of real life sentences normally found in news discourse. 
The sentences should resemble the provided example but should also vary. Like the provided example, the generated samples should not be overly simple. Each sentence pair should be separated with two newlines. \\ \\
Within each pair, the sentences should be separated with a single newline. Each sentence should start with 'A: ' or 'B: '. Apart from that do not generate any other output.
\end{mdframed}

\newpage
For factually neutral samples, the the following prompt was used: 

\begin{mdframed}
\mdfsubtitle{FACTUAL NEUTRALITY PROMPT}
'Factual neutrality' is a relationship between sentences A and B whereby has no effect on the likelihood of B being true.\\ \\
For example: \\
A: \{premise\} \\
B: \{hypothesis\} \\
\\ 
Generate 10 more pairs of sentences with a factual neutrality relationship.
The sentences should be diverse and reflect the type of real life sentences normally found in news discourse. 
The sentences should resemble the provided example but should also vary. Like the provided example, the generated samples should not be overly simple. Each sentence pair should be separated with two newlines. \\ \\
Within each pair, the sentences should be separated with a single newline. Each sentence should start with 'A: ' or 'B: '. Apart from that do not generate any other output.
\end{mdframed}

\section{Training Setup}
\label{app:training_setup}

The core dataset was randomly split to a training set (80\%) and a validation set (20\%). The core training set comprises 1205 samples, and the validation set comprises 302 samples. With the addition of the synthetically generated data and 500 pairs from the MNLI dataset, the training dataset comprises 12,249 sentence pairs.

Training was performed on an Nvidia A100 GPU, using Huggingface Transformers (v4.34.0) and PyTorch (v2.0.1). Fine-tuning was for 6 epochs, using early stopping on the validation loss. Best performing checkpoint on the validation set was kept. Otherwise, training used the default huggingface hyperparameters. GPT-3.5 was finetuned via the OpenAI API with the recommended default settings.

\section{Full Experimental Results}
\label{app:full_results}

\begin{table*}[htbp]
\renewcommand{\arraystretch}{1.2} 
\centering
\caption{Model results. Each entry indicates a single run.}
\label{tab:my_table}
\begin{tabular}{>{\centering\arraybackslash}p{2.0cm}ccccccc}
\hline
Gradient Training & Model & Data & Method & F1$_{\text{MAC}}$ & ACC \\
\hline
V & DeBERTa-large-NLI & Core + Synthetic & Focal Loss & \textbf{0.7} & 0.79 \\
V & DeBERTa-large-NLI & Core + Synthetic & Class Weights & 0.65 & 0.77 \\
V & DeBERTa-large-NLI & Core + Synthetic & Regular & 0.61 & 0.74 \\
V & DeBERTa-large-V3 & Core + Synthetic & Focal Loss & 0.37 & 0.58 \\
V & DeBERTa-large-V3 & Core + Synthetic & Class Weights & 0.61 & 0.75 \\
V & DeBERTa-large-V3 & Core + Synthetic & Regular & 0.28 & 0.71 \\
V & RoBERTa-base & Core + Synthetic & Focal Loss & 0.57 & 0.72 \\
V & RoBERTa-base & Core + Synthetic & Class Weights & 0.6 & 0.73 \\
V & RoBERTa-base & Core + Synthetic & Regular & 0.59 & 0.74 \\
V & DeBERTa-large-NLI & Core & Focal Loss & 0.68 & 0.8 \\
V & DeBERTa-large-NLI & Core & Class Weights & 0.66 & 0.75 \\
V & DeBERTa-large-NLI & Core & Regular & 0.67 & 0.78 \\
V & DeBERTa-large-V3 & Core & Focal Loss & 0.61 & 0.75 \\
V & DeBERTa-large-V3 & Core & Class Weights & 0.47 & 0.56 \\
V & DeBERTa-large-V3 & Core & Regular & 0.54 & 0.71 \\
V & RoBERTa-base & Core & Focal Loss & 0.4 & 0.7 \\
V & RoBERTa-base & Core & Class Weights & 0.45 & 0.61 \\
V & RoBERTa-base & Core & Regular & 0.41 & 0.68 \\
X & GPT-4 & None & Zero-Shot & 0.65 & 0.8 \\
X & GPT-4 & 3-shot & Few-shot & \textbf{0.7} & \textbf{0.81} \\
V & GPT-3.5 & Core & Regular & 0.69 & 0.78 \\
V & GPT-3.5 & Core + Synthetic & Regular & 0.63 & 0.77 \\
X & DeBERTa-large-NLI & None & No training & 0.54 & 0.72 \\
X & Baseline & Core & Cos. Sim. + DecisionTree & 0.38 & 0.61 \\ 
X & Baseline & Core + Synthetic & Cos. Sim. + DecisionTree & 0.44 & 0.63 \\ 
\hline
\end{tabular}
\end{table*}

\newpage
\newpage
\section{Zero-Shot and 3-shot Prompts}

For zero-shot classification with GPT-4, the following system prompt was used:

\begin{samepage}
\begin{mdframed}
\mdfsubtitle{SYSTEM PROMPT}
You are an advanced classifier.
\end{mdframed}
\end{samepage}

And the following instruction prompt:
\begin{samepage}
\begin{mdframed}
\mdfsubtitle{ZERO-SHOT CLASSIFICATION PROMPT}
You will classify the factual relationship between sentences A and B. The factual relationship can be either 'SUPPORTS', 'UNDERMINES', or 'NEUTRAL'. 'SUPPORTS' means that A factually supports B - if A is true, B is more plausible or likely to be true. 'UNDERMINES' means that A factually undermines B - if A is true, then B is less plausible or less likely to be true. 'NEUTRAL' means that the truthness of A has no implication on the likelihood of B being true. \\ \\
Here is a pair of sentences: \\ 
A: \{premise\} \\
B: \{hypothesis\} \\ \\ 
Classify their factual relation. Respond with 'SUPPORTS', 'UNDERMINES' or 'NEUTRAL', and nothing else.
\end{mdframed}
\end{samepage}

\newpage
For 3-shot classification, the same system prompt was used, in conjunction with the following instruction prompt:

\begin{samepage}
\begin{mdframed}
\mdfsubtitle{3-SHOT CLASSIFICATION PROMPT}
You will classify the factual relationship between sentences A and B. The factual relationship can be either 'SUPPORTS', 'UNDERMINES', or 'NEUTRAL'. 'SUPPORTS' means that A factually supports B - if A is true, B is more plausible or likely to be true. 'UNDERMINES' means that A factually undermines B - if A is true, then B is less plausible or less likely to be true. 'NEUTRAL' means that the truthness of A has no implication on the likelihood of B being true. \\ \\ 

Here's an example of two sentences with a 'NEUTRAL' relationship: \\
A: And with us having so much money invested into our honeymoon, we had no other choice but to board the ship. \\
B: The memory that will stick with her, she said, is when the ship stopped in Sri Lanka to refuel. \\ \\

Here are two sentences with a 'SUPPORTS' relationship: \\
A: Industry experts say the increase in milking cows has come from expansion of longstanding dairies, the launch of milking operations at existing farms that have diversified, and also from the relocation of dairy operations to South Dakota from states such as California. \\
B: As in other agricultural industries, dairy farmers are increasingly using genetics, data monitoring, technology and robotics to boost the production of each individual animal while implementing an economies-of-scale approach to the size of their farms, raising the efficiency and profitability of their operations. \\ \\
\end{mdframed}

\begin{mdframed}

And here are two sentences with an 'UNDERMINES' relationship: \\
A: Guinea had announced late Wednesday that it was canceling its participation to protect the health of its athletes. \\
B: North Korea is the only country to pull out of the Tokyo Olympics, also citing concerns related to COVID-19.
 \\ \\ 

Here is a new pair of sentences: \\
A: \{premise\} \\ 
B: \{hypothesis\} \\ \\

Classify their factual relation. Respond with 'SUPPORTS', 'UNDERMINES' or 'NEUTRAL', and nothing else.
\end{mdframed}
\end{samepage}

\end{document}